\documentclass{article}
\usepackage{spconf,amsmath,graphicx}

\usepackage{enumitem}
\setlist{nosep, leftmargin=14pt}

\usepackage{mwe} 

\title{Evaluating Automated Radiology Report Quality through Fine-Grained Phrasal Grounding of Clinical Findings}
\name{%
\begin{tabular}{@{}c@{}}
Razi Mahmood$^{1}$, 
Pingkun Yan$^{1}$,
Diego Machado Reyes$^{1}$,
Ge Wang$^{1}$\\ 
Mannudeep K. Kalra$^{2}$,
Parisa Kaviani$^{2}$, Joy T. Wu$^{3}$, Tanveer Syeda-Mahmood$^{3}$
\end{tabular}}

\address{
$^{1}$ Rensselaer Polytechnic Institute, Troy, NY USA., 
$^{2}$ Department of Radiology, \\
Massachusetts General Hospital, Harvard Medical School, Boston, USA.\\
$^{3}$ IBM Research, Almaden, San Jose, CA, USA }     

\begin{document}

\maketitle

\begin{abstract}
Several evaluation metrics have been developed recently to automatically assess the quality of generative AI reports for chest radiographs based only on textual information using lexical, semantic, or clinical named entity recognition methods. In this paper, we develop a new method of report quality evaluation by first extracting fine-grained finding patterns capturing the location, laterality, and severity of a large number of clinical findings. We then performed phrasal grounding to localize their associated anatomical regions on chest radiograph images. The textual and visual measures are then combined to rate the quality of the generated reports. We present results that compare this evaluation metric with other textual metrics on gold standard datasets.
\end{abstract}

\section{Introduction}
\begin{figure*}[t]
  \centering
  \centerline{\includegraphics[width=5.0in]{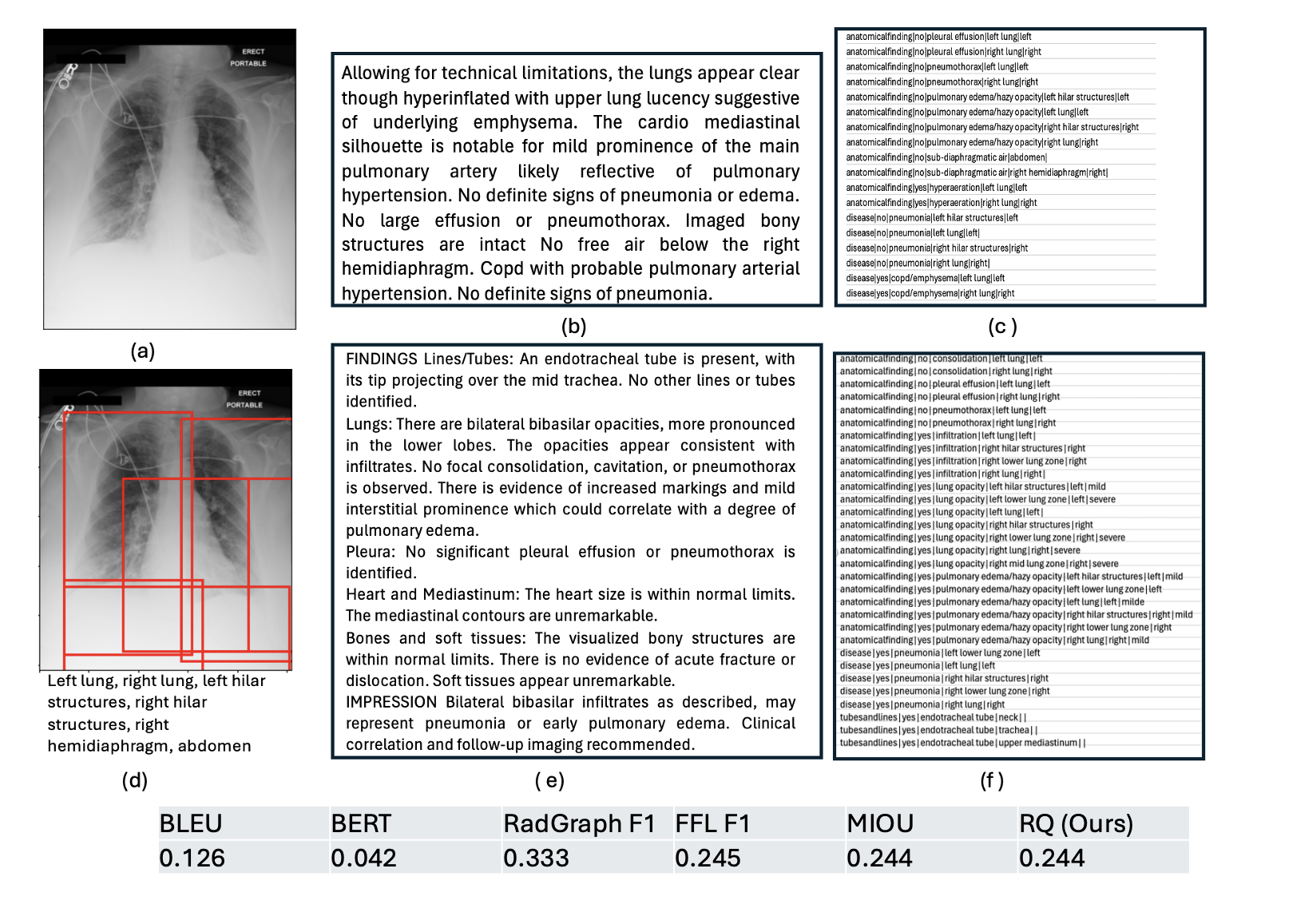}}
  \caption{(a) Illustration of the report quality problem. (a) Original image. (b) Ground truth report (Findings and Impressions only). (c) Fine-grained (FFL) patterns extracted from report of (b). (d) Anatomical locations of findings identified in (c) shown through bounding boxes. (e) Automated report produced by GPT-4. (f) FFL patterns extracted from automated report of (e). The table below shows the report evaluation scores produced by methods described in text for the automated report of (e).}\medskip
  \label{fig1}
\end{figure*}

With the evolution of AI models, it is now possible to produce realistic-looking natural language radiology reports, particularly for chest X-rays~\cite{Pang2023,syeda-mahmood2020,Endo2021,Nguyen2021}. 
Figure~\ref{fig1}e shows a sample report using  GPT-4 \cite{Ziegelmayer2023} on the chest X-ray image shown on Figure~\ref{fig1}a. While this appears good on surface, upon closer examination and comparing to the ground truth report shown in Figure~\ref{fig1}b, several mistakes can be found including the potential conclusion of pneumonia and missed pulmonary hypertension in the hilar regions. In general, the reports produced by generative AI tools can have false predictions, omissions, incorrect finding locations or  incorrect severity assessments.  

Identifying such factual errors, therefore, requires quality measures that can pay attention to both presence and absence of findings,  their locations, laterality and severity. Further, they should be robust to the different ways in which a finding is described.
Current methods for evaluating such descriptions  have so far been based on lexical or textual semantics-based scoring metrics\cite{bleuscore,Ziegelmayer2023,rouge,bertscore,radgraph,Liu2023quality,Yu2023}. While some metrics cover clinical entities and their relations\cite{radgraph,Yu2023}, generally scoring metrics do not explicitly capture the textual mention differences in the anatomy, laterality and severity. Further, phrasal grounding of the findings in terms of anatomical localization in images is not exploited in the quality scoring. 

In this paper, we propose a metric that captures both fine-grained textual descriptions of findings as well as their phrasal grounding information in terms of anatomical locations in images. We present results that compare this evaluation metric to other textual metrics on a gold standard dataset derived from MIMIC collection of chest X-rays and validated reports, to show its robustness and sensitivity to factual errors.
\vspace{-0.2in}
\section{Overall approach}
\label{sec:approach}
\begin{figure}[t]
\centering
  \centerline{\includegraphics[width=3.2in]{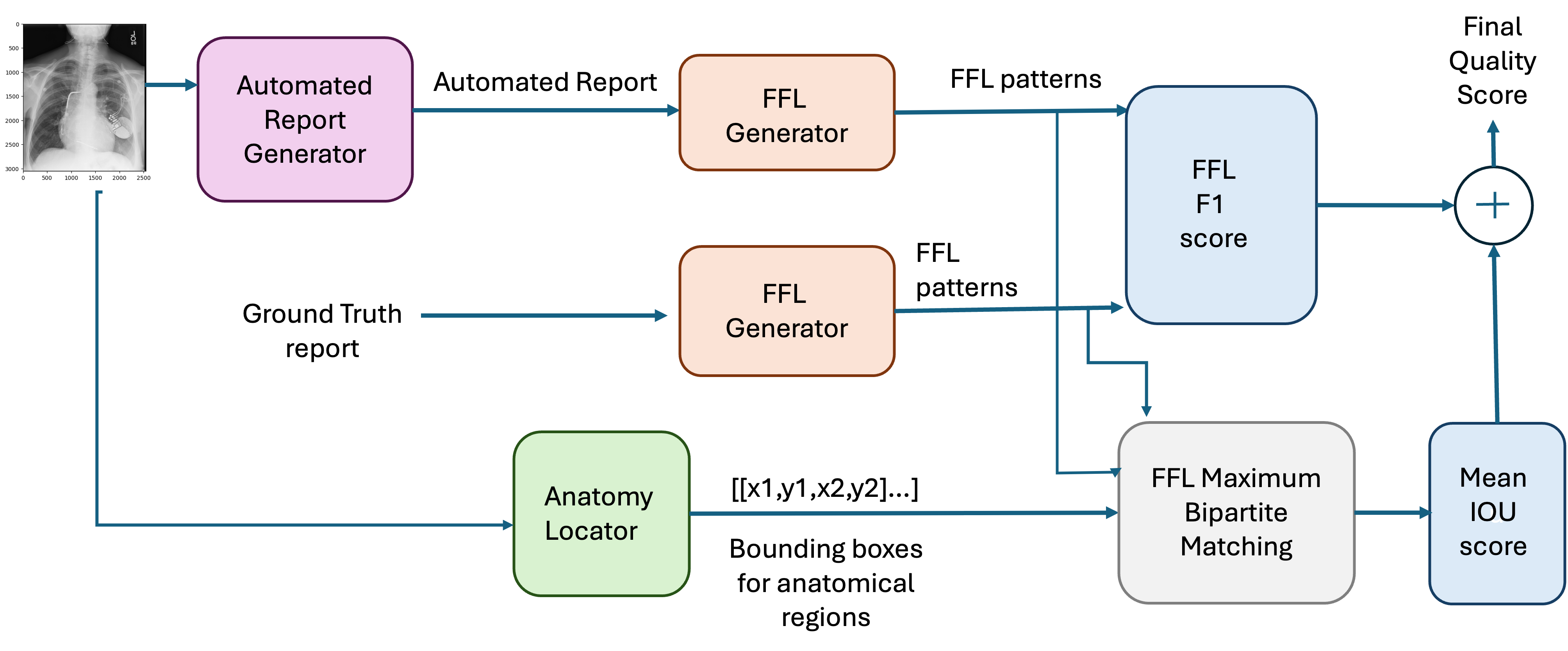}}
  \caption{Illustration of the overall approach to computing the report quality score.}\medskip
  \label{overview}
\end{figure}

Our overall approach to evaluating report quality is illustrated in Figure~\ref{overview}. Given a chest X-ray image and its associated ground truth report, we first extract fine-grained finding (FFL) patterns from the ground truth report as described in\cite{wu2020a}. This creates a structured description of the report using a normalized vocabulary for findings derived from a chest X-ray lexicon\cite{wu2020a}. Next, we extract all important anatomical region bounding boxes as defined in the ChestImagenome dataset\cite{Wu2021} and assign them to the relevant findings based on the anatomical location mentioned in the FFL textual pattern.  Next, to evaluate an automated report generated for the same image by available methods\cite{Pang2023,syeda-mahmood2020,xraygpt,rgrg}, we similarly extract FFL patterns from the automated report. The overlap in FFL patterns of automated and ground truth reports is evaluated in terms of precision, recall and F1-score. A geometric comparison is then initiated with the pair of FFL patterns from the ground truth and automated regions using the bounding boxes of the referred anatomical locations within each FFL to do phrasal grounding of the underlying findings. The spatial overlap of the anatomical regions indicated in the findings constitutes a geometric similarity measure per FFL pattern. A bipartite graph is formed from the FFL pairs using the spatial overlap measure as edge weights.  The mean IOU score derived from the maximum matching in the bi-partite graph serves as the phrasal grounding metric. The final quality score is then the average of the F1 and mean IOU scores, reflecting a match in both the textual description of findings and their implied anatomical locations. 

\noindent{\bf Extraction of FFL patterns}
\label{fflextraction}

A fine-grained finding pattern (FFL) $F_{i}$ describes a finding in terms of its presence or absence and modifiers as described in\cite{syeda-mahmood2020}. For our report evaluation purpose, we restrict the modifiers to cover  location, laterality and severity of a finding so that an FFL is denoted by:
\begin{equation}
    F_{i}= T_{i}|N_{i}|C_{i}|A_{i}|L_{i}|S_{i}
    \label{ffl}
\end{equation}
\noindent where $T_{i}$ is the finding type, $N_{i}= yes|no$ indicates a present or absent finding respectively, $C_{i}$ is the normalized core finding name, $A_{i}$ is the anatomical location, $L_{i}$ reflects laterality, and $S_{i}$ reflects the severity of the core finding $C_{i}$. The normalized values for each category of information captured in the FFL patterns was derived from a comprehensive clinician-curated chest X-ray lexicon described in\cite{wu2020a,syeda-mahmood2020a}.

To extract the FFL patterns, adopting a vocabulary driven approach described in earlier work\cite{syeda-mahmood2020,syeda-mahmood2020a}, we detect the core finding and their modifiers using the chest X-ray lexicon. We then group noun phrases and apply negation detection and pattern completion as described in \cite{syeda-mahmood2020,syeda-mahmood2020a}.  
The FFL pattern completion step uses a priori domain knowledge to incorporate anatomical locations.  For example, ”alveoli” would be inserted for an alveolar finding, even when not specified in the report sentence. Thus the final FFL patterns show more information than the original sentence. Further,  due to the normalization of names through the lexicon, they are robust to variations in descriptions of the same finding across reports.  

Figures~\ref{fig1}c and f show the FFL patterns derived from the sentences in the ground truth and automated reports from Figures~\ref{fig1}b and e respectively. As can be seen, FFL labels capture the essence of a report adequately and much more comprehensively than core findings alone.  The FFL label extraction algorithm reported in \cite{syeda-mahmood2020} is known to be highly accurate in terms
of the coverage of findings with around 3\% error mostly due to negation sense
detection. 

\begin{table*}
\caption{Evaluation of Report Quality- Lexical metrics (FFL F1-score).}\label{qualityscore}
\centering
\begin{tabular}{l|l|l|l|l|l}
\hline
{\bf Report Origin} & {\bf Extent of FFL patterns} & {\bf Avg. Precision} & {\bf Avg. Recall} & {\bf Avg. F1-score} & {\bf Avg. MIOU}\\
\hline
{\bf RGRG} &  All & {\bf 0.313} & {\bf 0.399} & {\bf 0.331} & 0.487 \\
RGRG & Anatomy & 0.440 & 0.598 & 0.486 & 0.487\\
RGRG & Core Finding & 0.453 & 0.643 & 0.511 & -\\
\hline
X-RayGPT & All & 0.256 & 0.234 & 0.223 & 0.357\\
X-RayGPT & Anatomy  & 0.391 & 0.377 & 0.357 & 0.357 \\
X-RayGPT & Core Finding & 0.430 & 0.405 & 0.392 & - \\
\hline
GPT4 & All & 0.216 & 0.327 & 0.242 & 0.368 \\
GPT4 & Anatomy  & 0.326 & 0.481 & 0.367 & 0.368\\
GPT4 & Core Finding & 0.330 & 0.524 & 0.388 & - \\
\hline
\end{tabular}
\end{table*}
\begin{table*}
\caption{Evaluation of Report Quality-All Metrics.}\label{comparison}
\centering
\begin{tabular}{l|l|l|l|l|l|l|l}
\hline
{\bf Report Origin} & {\bf \# Reports} & {\bf Avg. FFL F1-score} & {\bf Avg. IOU score} & {\bf Combined} & {\bf BLEU} & {\bf BERT} & {\bf RadGraph F1}\\
\hline
{\bf RGRG} &{\bf 439} & {\bf 0.440} & {\bf 0.487} & {\bf 0.463} & {\bf 0.237} & {\bf 0.329} &{\bf 0.529}\\  
XrayGPT & 439 & 0.391 & 0.357& 0.374& 0.145 & 0.256 & 0.390\\
GPT4 & 439 & 0.326 &0.368 & 0.347 & 0.106 & 0.087 &0.434\\
\hline
\end{tabular}
\end{table*}
\noindent{\bf Extraction of anatomical locations from images}
\label{geometry}

To extract the anatomical locations from images, we adopted the approach described in \cite{Wu2020,Wu2021} that predicts bounding boxes corresponding to a list of 36 anatomical regions cataloged in the chest X-ray lexicon and provided in the ChestImagenome dataset\cite{Wu2021}. The bounding boxes for the anatomical regions were detected using a Faster RCNN model trained on labeled bounding boxes in chest X-ray images\cite{Wu2021}. 
Figure~\ref{fig1}d shows bounding boxes of the anatomical regions identified in the FFL patterns of the ground truth radiology report in Figure~\ref{fig1}c covering anatomical locations of right and left lung, hilar structures, abdomen, and hemidiaphragm. The localization accuracy of the bounding box detector was previously assessed at 0.896 precision and 0.881 recall and was used to reliably generate the ChestImagenome benchmark dataset\cite{Wu2021}.
\vspace{-0.1in}
\section{Developing report evaluation score}
\label{overallscore}

We now describe our clinical accuracy score using the structured representation of the findings in terms of FFL patterns and their phrasal grounding.  Specifically, given a ground truth radiology report G and a predicted automated report P, we extract FFL pattern set from sentences within these reports as $F_{G}$ and $F_{P}$ from Equation~\ref{ffl}. For each FFL pattern $F_{i}$, we also form prefix patterns obtained by successively removing modifier descriptions as:
\begin{equation}
    W_{j}(F_{i})=T_{i}|N_{i}|C_{i}|M_{1}|...|M_{j}
\end{equation}
\noindent with $M_{j}$ is the jth modifier retained in the FFL pattern. By creating prefixes of patterns at modifier boundaries, we can assess the quality of matching at various levels of granularity.  

\noindent{\bf FFL F-1 score}
\label{f1scoreeval}

Given two prefix versions of FFL patterns between ground truth report and generated automated report, we can calculate the true positives ($t_{p}$), and false positives ($f_{p}$)and false negatives ($f_{n}$) to computer F1 score as:
\begin{eqnarray}
    t_{p}=|W_{j}(F_{Gi})|, \mbox{ s.t.} W_{j}(F_{Gi})=W_{j}(F_{Pk})\\
    f_{p}=|W_{j}(F_{Pk})|-t_{pj}, \mbox{   }
    f_{n}=|W_{j}(F_{Gi})|-t_{pj}\\
    F1_{G,P}=\frac{2t_{p}}{2t_{p}+f_{p}+f_{n}}
\end{eqnarray}
\noindent Here $W_{j}(F_{Gi}$ and $W_{j}(F_{Pk}$ are the matching FFL patterns from ground truth and automated report respectively.

\noindent{\bf MIOU score}
\label{mioueval}

To evaluate the geometric overlap between the findings, we consider FFL patterns that indicate the same core finding prefix (i.e. match in $T_{i}|N_{i}|C_{i}$). Since the same core finding can be observed in multiple locations (e.g. left upper lobe, and right lower lobe), several possible matches exist between pairs of FFL patterns of ground truth and generated reports. To compute the overlap between the indicated spatial locations in the pairs, we use the IOU score. Specifically, let the anatomical location bounding box in an FFL pattern $F_{Pk}$  of a predicted report be denoted by $B_{Pk}=<x1,y1,x2,y2>$, and let $B_{Gi}=<x_{g}1,y_{g1},x_{g2},y_{2}>$ be the anatomical location of the corresponding ground truth finding. Then the IOU score is given by:
\begin{equation}
I_{ki}=\sum_{i}\frac{|B_{Pk}\cap B_{Gi}|}{|B_{Pk}\cup B_{Gi}|}
\end{equation}

To find the best pair of corresponding findings, we treat the FFL patterns of ground truth report and automated report as a bipartite graph and perform a maximum matching using the IOU score to weigh the edges. 
The resulting cost of the maximum matching is then given by $I_{GP}=\sum_{i}I_{GiPj}$,  for corresponding $F_{Pj}$ and $F_{Gi}$. The mean IOU score per pair of reports per image is then given as 
\begin{equation}
MIOU(G,P)=\frac{2*I_{GP}}{|F_{G}|+ |F_{P}|}
\end{equation}
Combining the lexical and geometric aspects of the match, we form an overall report quality score per image as:
\begin{eqnarray}
    RQ(G,P)=F1(G,P)+MIOU(G,P)\\
    RQ=\frac{\sum_{G,P}RQ(G,P)}{|G|}
\end{eqnarray}
\noindent where $RQ(G,P)$ is per pair of ground truth and automated reports, and  $|G|=|P|$ represent the image collection over which the pairs of reports are analyzed for assessment.

\begin{tiny}
\begin{table}
\caption{Sensitivity to factual error perturbations. }\label{sensitivity}
\centering
\begin{tabular}{l|l|l|l|l}
\hline
{\bf Method} & {\bf Reports} & {\bf Finding } & {\bf Location}  & {\bf Severity}\\
\hline
BLEU& 500 & 0.3  & 0.1  & 0.1\\  
BERT & 500 & 0.2 & 0.14& 0.09\\
Radgraph F1 & 500 & 0.3 &0.15  & 0.09\\
RQ(Ours) & 500 & 0.5 & 0.4 & 0.39\\
\hline
\end{tabular}
\end{table}
\end{tiny}
\section{Results}
We now present results of applying the quality score to assess report quality on a benchmark dataset of chest X-ray images with validated ground truth reports. 

\noindent{\bf Dataset}: For our experiments, we selected the gold dataset of 439 chest x-rays and their ground truth reports from the publicly available clinician validated ChestImagenome\cite{Wu2021} collection built from the MIMIC dataset\cite{mimic-4} and vetted for bias and fairness during their IRB approval. 
The dataset also provided  FFL patterns covering 60 findings extracted from the finding and impression sections of the ground truth reports to serve for our report quality evaluation.  Further, 36 anatomical locations were marked in each of the images and validated by 2 clinicians.  To evaluate report quality, we experimented with open source radiology report generation tools, namely, XrayGPT\cite{xraygpt} and RGRG\cite{rgrg}, and an internal tool based on GPT-4 being piloted in our hospital. We ran the report generation tool on the benchmark dataset and retained their automated reports. We then extracted the FFL patterns and recorded the bounding boxes of their indicated anatomical locations for the computation of the report quality score. 


\noindent{\bf Evaluation through the proposed measure}: 

Next, we evaluated the report quality using our lexical measure reported in Section~\ref{f1scoreeval} using prefix patterns restricting to core finding, anatomy (with laterality),  severity. The result is shown for the 3 report generators evaluated in Table~\ref{qualityscore}. From this table, we observe that the RGRG report generator has the highest lexical quality with their FFL patterns matching closely with the ground- truth FFL patterns at all levels of granularity. We also notice that all methods improved in report quality when evaluated on the basis of their core finding. The Mean IOU scores were evaluated for the FFL prefixes that retained the anatomical location and are as shown in the last column of that table.  

\noindent{\bf Comparison with evaluation scores}:

To compare our approach with other report evaluation scores, we selected representative methods for word overlap scores (BLEU\cite{bleuscore}), semantic textual matching (BERTscore\cite{bertscore}) and clinical accuracy F1-score \cite{radgraph}.  The result is shown in Table~\ref{comparison} from which we observed that the lexical comparison scores under-estimated the accuracy of the reports due to lexical mismatch in the reported descriptions.  The clinical accuracy score, as it was trained on fewer findings (14 findings), overestimated the performance by giving higher scores due to missed findings in their model. Our approach gave balanced estimates of report quality indicating 36-48\% spatial overlap of their locations and 33-44\% overlap in their descriptions. Finally, we observe that all reporting metrics rated the RGRG report as the best even in this evaluation. 

\noindent{\bf Sensitivity of the report quality score}: 

To measure sensitivity, we created 500 additional synthetic reports by perturbing each ground truth reports to introduce a range of errors in findings in terms of negation reversal, substitutions,  and alteration in location and severity . 
The FFL pattern extraction, and spatial localization of findings was completed on the synthetic reports and all quality scores were re-evaluated by comparing the synthetic reports to the associated ground truth reports. The interval change of scores was taken as a measure of sensitivity of the report evaluation score to the factual errors. The result is shown in Table~\ref{sensitivity} for all report evaluation measures. As can be seen, the lexical and semantic score changes remained generally low, while the Radgraph clinical accuracy F1 score showed less sensitivity to location variations. In comparison, our quality score showed good range of variation to reflect quality in such fine-grained characterization.

\noindent{\bf Discussion}: Our method exploited the notation of standardized locations for anomalies in chest X-rays. While the metric itself which focuses on both identity and location differences could be used to evaluate phrasal grounding in general scenes, the FFL pattern descriptions may not have such standardized locations. Finally, our method is computationally efficient even for the bipartite matching step due to the small number of regions matched. 

\section{Conclusions}
\label{sec:illust}
In this paper, we present a new approach to evaluating the quality of generated chest X-ray radiology reports. Our approach captured fine-grained finding patterns along with phrasal grounding of findings and is shown to be sensitive to factual errors in radiology reports making it suitable as an evaluation metric for fact-checking of radiology reports.


\bibliographystyle{IEEEbib}
\bibliography{refs}

\end{document}